\documentclass[conference]{IEEEtran}
\usepackage{algorithm}
\usepackage{algorithmic}
\usepackage{multirow}
\usepackage{multicol}
\usepackage{times}
\usepackage{latexsym}
\usepackage{graphicx}
\usepackage{graphicx}
\usepackage{algorithm}
\usepackage{algorithmic}
\usepackage{multirow}
\usepackage{array}
\usepackage{mdwmath}
\usepackage{mdwtab}
\usepackage{url}
\usepackage{soul}
\usepackage{longtable}
\usepackage{epsfig}
\usepackage{amssymb}
\usepackage{amsmath}
\usepackage{amsfonts}
\usepackage{breqn}
\usepackage{algorithm}
\usepackage{color}

\usepackage{multirow}

\usepackage{amsmath}

\newlength\myindent
\newcommand\bindent{%
  \begingroup
  \setlength{\itemindent}{\myindent}
  \addtolength{\algorithmicindent}{\myindent}
}
\newcommand\eindent{\endgroup}
\newif\ifcomment
\def\cb{\ifcomment \textcolor{blue}}

\newif\ifextra
\def\extra{\ifextra \textcolor{green}}
\ifCLASSINFOpdf
\else
\fi
\begin{document}
%
\title{An Iterative Transfer Learning Based Ensemble Technique for Automatic Short Answer Grading}
\author{\IEEEauthorblockN{Shourya Roy}
\IEEEauthorblockA{Xerox Research Centre India\\
Bangalore, India\\
Email: shourya.roy@xerox.com}
\and
\IEEEauthorblockN{Himanshu S. Bhatt}
\IEEEauthorblockA{Xerox Research Centre India\\
Bangalore, India\\
Email: himanshu.bhatt@xerox.com}
\and
\IEEEauthorblockN{Y. Narahari}
\IEEEauthorblockA{Indian Institute of Science\\
Bangalore, India \\
Email: hari@csa.iisc.ernet.in}}


%


\maketitle

\begin{abstract}
Automatic short answer grading (ASAG) techniques are designed to automatically assess  \textit{short} answers to questions in natural language, having a length of a few words to a few sentences. Supervised ASAG techniques have been demonstrated to be effective but suffer from a couple of key practical limitations. They are greatly reliant on instructor provided model answers and need labeled training data in the form of graded student answers for every assessment task. To overcome these, in this paper, we introduce an ASAG technique with two novel features. We propose an iterative technique on an ensemble of (a) a text classifier of student answers and (b) a classifier using numeric features derived from various similarity measures with respect to model answers. Second, we employ  canonical correlation analysis based transfer learning on a common feature representation to build the classifier ensemble for questions having \textit{no} labelled data. The proposed technique handsomely beats all winning supervised entries on the S\textsc{ci}E\textsc{nts}B\textsc{ank} dataset from the ``Student Response Analysis'' task of SemEval 2013.  Additionally, we demonstrate generalizability and benefits of the proposed technique through evaluation on multiple ASAG datasets from different subject topics and standards. 
\end{abstract}


%
\IEEEpeerreviewmaketitle

\section{Introduction}
Computer assisted assessment has been prevalent in schools and colleges for many years albeit primarily for questions with constrained answers. To answer such \textit{recognition} questions e.g. multiple choice questions (MCQs), students typically  have to choose the correct answer(s) from a given list of options. \extra{While computer aided assessment (typically on a binary scale correct/incorrect) in this scenario is almost trivial these days,}\fi Prior work reported in many papers has questioned the effectiveness of such questions to assess knowledge, scholarship, and depth of understanding  gathered by students~\cite{conole2005review, martinez1992review\extra{,wang2008assessing}\fi}.
\extra{Firstly, they are less reliable owing to the possibility of pure guessing paying some dividends. Techniques which do not account for influence of guessing strategies used by students do not lead to reliable assessment~\cite{singley1995open} Secondly, the presence of alternative responses provide inadvertent hints which may change the very
nature of problem-solving and reasoning\cite{conole2005review,hirschman2000automated}. Finally, in many cases, MCQs are not appropriate to measure acquired knowledge such as hypothetical reasoning and self-explanation in Science courses~\cite{conole2005review, martinez1992review,wang2008assessing}}\fi 
 On the other hand, open-ended questions or \textit{recall} questions that seek  constructed responses from students, reveal their ability to integrate, synthesize, design, and communicate ideas in natural language.  An example is shown in Table~\ref{tab:summaryfeat}. However, automatic assessment of such answers on various scales (binary, ordinal, nominal) has remained a non-trivial challenge owing to multiple reasons. These include linguistic variations of student answers to a question (same answer could be articulated in  different ways); lack of uniformity in how instructors provide model answers across questions and datasets (detailed, brief, representative); subjective nature of assessment (multiple possible correct answers or no correct answer); lack of consistency in human rating, etc.  Consequently, the task of assessment of answers to recall questions has predominantly remained a repetitive and tedious manual job for teaching instructors\extra{and is often seen as an overhead and non-rewarding by them}\fi. This paper dwells on a computational technique for automatically grading such answers and particularly focuses on {\it short answers} which are a few words to a few sentences long (everything in between fill-in-the-gap and essay type answers~\cite{burrows2015eras}). This task of automatically grading  short answers is referred to as \textit{Automatic Short Answer Grading (ASAG)}. 

\begin{table*}
\centering
\small
\begin{tabular}{p{1.0cm}|p{8.0cm}|p{8.0cm}}
\hline
\textbf{Question} &  \textbf{(Q1)} What are the main advantages associated with object-oriented programming? (5) & \textbf{(Q2)} How are overloaded functions differentiated by a compiler? (5)\\
\hline
\textbf{Model Ans} &   Abstraction and reusability. & Based on the function signature. When an overloaded function is called, the compiler will find the function whose signature is closest to the  given function call.\\
\hline
\textbf{Ans-1} &   This type of programming is more flexible, making it easier to add and modify the program.  It is also a type of a fail safe program, you check each individual module.  This eliminates redundant code and makes the program easier to read for other programmers.  When debugging the program it is easier to track down the source of a problem within a module rather than a 2 million line program. (5/5) & it looks at the number, types, and order of arguments in the function call.(5/5)\\
\hline
\textbf{Ans-2} &   The advantage is that OOP allows us to build classes of objects.  Three principles that make up OOP are:  Encapsulation- Objects combine data and operations.  Inheritance- Classes can inherit properties from other classes.  Polymorphism- Objects can determine appropriate operations at execution time. (2.5/5) & they have to have same return type, but different input parameters (3/5)\\
\hline \\
\end{tabular}
\caption{Examples of questions, model answers, and student answers with instructor given scores from an undergraduate computer science course \cite{mohler2009text}}.
\label{tab:summaryfeat}
\vspace{-35pt}
\end{table*}

A large fraction of prior work in ASAG  has been based on supervised learning techniques viz.  classification and regression. These techniques extract various features from model and instructor graded student answers using natural language processing (NLP) techniques reflecting \textit{similarity} (synonymously, \textit{overlap}, \textit{correspondence}, \textit{entailment} etc.) between them. 
For example, Dzikovska et. al. proposed four lexical  similarity metrics viz. the raw number of
overlapping words, F1 score, Lesk score and cosine
score between student and model answers as features \cite{dzikovska2012towards}. These features are then fed to various classification or regression techniques to train models which can subsequently be applied to score new student answers automatically. While classification techniques can predict scores directly, continuous valued regression output needs to be discretized based on some thresholding  logic (e.g. \textit{ceil}, \textit{floor}). Such supervised techniques trained solely on features derived with respect to model answers immediately suffer from a couple of intuitive shortcomings. Firstly, the nature of model answers varies across questions. Consider the two examples shown in Table~\ref{tab:summaryfeat}: the first question has a very brief model answer compared to the second one. Consequently, the same type of features may not be able to effectively measure similarity of student answers with respective model answers for both questions. Secondly, student answers can be (very) different from the corresponding model answers but could still be correct. Consider an example-seeking question: \texttt{\small Give an example of a Java primitive Wrapper class.} The model answer may not exhaustively list all possible answers (in fact, it may be impossible in some cases) e.g. \texttt{\small Byte, Short, Integer, Long, Float etc.} students may write. \extra{To train these models, instructors need to grade a fraction of student answers and provide as {\it training data}. The remaining student answers as well as new answers (\textit{test data}) can be automatically graded by applying these models.}\fi 

\subsection*{Ensemble Based Supervised ASAG}
To address the above  shortcomings (which leads to the first contribution of this work), we propose a novel supervised ASAG technique based on an ensemble of two classifiers. The \textit{first classifier} is a text classifier trained using the classical TFIDF representation \cite{salton88} of bag-of-words (BoW) features of student answers. It is independent of model answers and learns good textual features (words and n-grams) from graded student answers to discriminate between student answers belonging to different scores. The \textit{second classifier} has features expressed as real numbers indicating similarity of student answers with the corresponding model answer (analogous to model answer based classifiers). We take various lexical, semantic, and vector space based measures to compute these features (a.k.a similarity values). The classifiers complement each other splendidly since the first  (text based) classifier is independent of the model answer whereas the second classifier is based on the similarity between the model answer and the student answers. By exploiting student answers directly in the first classifier, additionally, the ensemble can overcome the shortcomings mentioned earlier in this paragraph. While stacking of classifiers has been used in ASAG \cite{sakaguchi-heilman-madnani:2015:NAACL-HLT} towards ``one-shot'' combination of predictions from multiple classifiers, the proposed technique is designed in a different (iterative) manner to eliminate the need for extensive labelled data for new questions (will be explained next). We empirically demonstrate that this iterative ensemble outperforms (significantly, in many cases) either of the constituent classifiers (\textbf{\S}~\ref{sec:detail_results}).

\subsection*{Transfer Learning for ASAG}
While supervised models have been applied in many real-life scenarios to automate human activities, we opine that ASAG does not readily fit into the same {\it train-once-and-apply-forever} model.  Every assessment task is unique and hence, graded answers from one assessment task cannot readily be used to train a model for another. In today's world, repetition of questions across different groups of students is a rarity owing to proliferation of sharing and communication channels.  Consequently, application of supervised ASAG techniques would require ongoing instructor involvement to create labelled data (by grading $\frac{1}{2}$ to $\frac{2}{3}^{rd}$ of student answers as per typical train-test split guidelines) for every question and assessment task. Requirement of such continuous involvement of instructors limits the benefit of automation and thereby poses a hindrance to practical adoption. Towards addressing this limitation, we  propose a \textit{transfer learning} based approach for ASAG.

Transfer learning techniques, in contrast to traditional supervised techniques, work on the principle of transferring learned knowledge across \textit{domains}. In transfer learning parlance, a domain $D$ consists of two components: an $n$-dimensional feature space $\mathcal{X}$ and a marginal probability distribution $P(X)$ where $X=\{x_1,x_2,\ldots,x_n\} \in \mathcal{X}$. Two domains, commonly referred to as \textit{source} (with labeled data; typically aplenty) and \textit{target} (with less or no labelled data), are said to be different if they have different feature spaces or different marginal probability distributions \cite{tr_survey}. Supervised models trained on data from the source domain cannot be applied to the target domain data as it would violate the fundamental assumptions that training and test data must be in the same feature space and have the same distribution. In such cases, transfer learning techniques have been shown to be effective (to reduce new labeling efforts) for various tasks such as sentiment classification \cite{SCL1}, named entity recognition (NER) \cite{daume2009frustratingly} and social media analytics \cite{bhatt-conll}. Surprisingly, we found only one prior work \cite{heilman2013ets} where domain adaptation was used for ASAG based on the technique proposed in \cite{daume2009frustratingly}. We will review this along with other prior work in Section~\ref{sec:prior:tlasag} and empirically compare against in Section~\ref{sec:agg_results}.

As the second contribution of this work, we employ a novel transfer learning based technique for ASAG by considering answers to different questions as different domains\extra{as they have different marginal probability distributions of features and possibly different features too}\fi. The technique leverages canonical correlation analysis (CCA) for building the classifier ensemble for target questions requiring graded answers only from the source question; this eliminates the need for graded answers for the former. It transfers the trained model of the second classifier (model answer based) of source question ensemble by learning a common shared representation of features which minimizes domain divergence and misclassification error. The transferred model is applied on answers to target questions to predict scores and confidently predicted answers are considered as \textit{pseudo labelled data} to train the corresponding first classifier. This, along with the transferred second classifier, constitutes the ensemble for the target question. The ensemble is then applied to the remaining (other than the pseudo labelled data) student answers. In an analogous manner, confidently predicted instances from the ensemble are added to the pseudo labelled data pool to update the first classifier. The ensemble is then iteratively applied and used to augment the pseudo labelled data pool till all answers are confidently classified or some predefined stopping criteria is met. It is imperative to note that we do not require \textit{any} instructor graded student answers for the target question in this entire iterative process. Secondly, a similar transfer would have been less meaningful to be applied to the first (text) classifier. Between answers to the two questions, the feature space and distributions of features are both expected to be different. For example, the perfect scoring student answer of (Q1) in Table~\ref{tab:summaryfeat} would be a totally incorrect answer for (Q2). 

\subsection*{Contributions}
We propose a novel supervised ASAG technique using an ensemble of a text and a numerical classifier (\textbf{\S}~\ref{sec:approach}). We introduce a transfer learning technique for ASAG towards reducing continuous labeling effort needed for the task, thus taking a step towards making supervised ASAG techniques practical (\textbf{\S}~\ref{sec:approach}).

We empirically demonstrate superior performance of the proposed method in comparison to \cite{heilman2013ets} on the dataset released by \cite{dzikovska2013semeval} for the joint task of student response analysis in SemEval 2013 Task 7.  Additionally, we provide a detailed quantitative  analysis on the dataset collected as a part of an undergraduate computer science course \cite{mohler2011learning} towards bringing out insights on why and when transfer learning in ASAG produces superior performance. (\textbf{\S}~\ref{sec:eval})

We believe that this is one of the first efforts in ASAG which reports empirical results of a technique across multiple datasets towards filling an important gap in this field. \footnote{Quoting from a recent survey paper \cite{burrows2015eras}, ``[Finally, concerning the effectiveness scores in Table 7,] the meaningful comparisons
that can be performed are limited, as the majority of evaluations have been performed
in a bubble. That is, the data sets that are common between two or more publications
are relatively few.''} (\textbf{\S}~\ref{sec:eval})

\extra{

In the next section, we provide a detailed review of relevant prior art from supervised ASAG and transfer learning topics. The following section ($\S$~\ref{sec:approach}) is the heart of the paper describing our  approach in detail - explaining the intuition first followed by formal technical description. Section~\ref{sec:eval} provides details of empirical evaluation along with datasets and metrics used. The final section ($\S$~\ref{sec:conc}) concludes the paper and lists possible future research directions.
}\fi

\section{Prior Art}
\label{sec:prior}
Two recently written survey papers by Burrows et. al. \cite{burrows2015eras} and Roy et. al. \cite{roy2015perspective} provide comprehensive views of prior research in ASAG. In this section, we review relevant topics for our technique viz. supervised ASAG, transfer learning and transfer learning for ASAG .

\subsection{Supervised ASAG}
\label{sec:prior:supasag}
Most prior work in supervised ASAG took the approach of designing novel task and dataset specific features to feed to standard classification and regression algorithms. Sukkarieh used features  based on lexical constructs such as presence/absence of concepts, the order in which concepts appear etc.~\cite{sukkarieh2011using,sukkarieh2009c}; 
CAM (Content Assessment Module)  used types of overlap including word unigrams and n-grams, noun-phrase chunks, parts of speech~\cite{Bailey:2008:DME:1631836.1631849}; Madnani et. al. applied BLEU score (commonly used for evaluating machine translation systems), ROUGE (a recall based metric that measures the lexical and phrasal overlap between two pieces of text) for summary assessment~\cite{madnani2013automated} and Nielsen et. al. used carefully crafted lexical and syntactic features~\cite{nielsen2008taxonomy}. Horbach et. al. demonstrated an interesting variation for assessment of reading comprehension questions where they used the original reading text as a feature~\cite{horbach2013using}. Sukkarieh et. al. compared different classification techniques such as k-Nearest Neighbor, Inductive Logic Programming, Decision Tree and Na\"{i}ve Bayes to compare two sets of experiments viz. on raw text answers and annotated answers ~\cite{Pulman:2005:ASA:1609829.1609831,sukkarieh2005information}.

\extra{{\bf Regression based techniques}: Regression techniques were used for automated assessment to arrive at a real valued score which are then discretized possibly as per scoring scale. Here again we see prevalence of focus on feature design with standard regression techniques.  The e-Examiner system  used linear regression on ROUGE metrics~\cite{gutl2008moving}, Sil et. al. used Support Vector Machines with Radial Basis Function kernels and Wang et.al. applied a regression technique for assessing creative answer assessment~\cite{wang2008assessing}.}\fi

\extra{Sil et. al. used Support Vector Machines with Radial Basis Function kernels (RBF-SVM) for learning non-linear regression models of grading with several higher order features derived from free-text answers~\cite{sil2012automatic}. The e-Examiner system  uses linear regression on ROUGE metrics~\cite{gutl2008moving}. Wang et.al. applied a regression technique for assessing creative answer assessment~\cite{wang2008assessing}. }\fi

\extra{Sukkarieh used the Maximum Entropy classifier using features  based on lexical constructs such as presence/absence of concepts, the order in which concepts appear, role of a word in a sentence (e.g. active/passive) etc. to predict if a student response is entailed in at least one of the model answers~\cite{sukkarieh2011using}. This technique built on the prior work by Sukkarieh and others on concept mapping based ASAG by extracting higher level features indicating clause structure, negation etc.~\cite{sukkarieh2009c}.  Earlier, they compared different classification techniques such as k-Nearest Neighbor, Inductive Logic Programming, Decision Tree and Na\"{i}ve Bayes to compare two sets of experiments viz. on raw text answers and annotated answers ~\cite{Pulman:2005:ASA:1609829.1609831,sukkarieh2005information}.
Nielsen et. al. used carefully crafted features using NLP pre-processing obtained from lexical and syntactic forms of student answers~\cite{nielsen2008taxonomy}. Various lexical similarity scores such as number of overlapping words, $F_1$ score, Lesk score and cosine score were used in conjunction with classical decision tree classifier to train a Decision Tree classifier to categorize student responses into one of the 5 categories~\cite{dzikovska2012towards}. CAM (Content Assessment Module)  used a
k-nearest neighbor classifier and features that measure the percentage overlap of content
on various linguistic levels between the teacher and student answers~\cite{Bailey:2008:DME:1631836.1631849}. The types
of overlap include word unigrams and trigrams, noun-phrase chunks, text similarity thresholds, parts of speech, lemmas, and synonyms. For summary assessment, Madnani et. al. used a logistic regression classifier on a 5 point scale~\cite{madnani2013automated}. They used interesting features to capture commonalities between an original passage and a summary such as BLEU score (commonly used for evaluating Machine Translation systems), ROUGE (a recall based metric that measures the lexical and phrasal overlap between two pieces of text), overlap of words and phrases etc. Horbach et. al. demonstrated an interesting variation for assessment of reading comprehension questions where they used the original reading text as a feature~\cite{horbach2013using}.}\fi

Dzikovska et. al. \cite{dzikovska2013semeval} floated a task, ``Student Response Analysis'' (SRA) in the Semantic Evaluation (SemEval) workshop in 2013, where participating teams had to categorize student answers into 2-,3- and 5-way categorization.\footnote{\url{https://www.cs.york.ac.uk/semeval-2013/task7/}} As a part of this task, they also released a dataset of student answers split into train and test format. This is one of the few well annotated datasets in ASAG though for the 5-way categorization it has an atypical characteristic of \textit{nominal} grades (labels) viz. `Correct', `Partially correct incomplete',`Contradictory' (student answer contradicts the model answer), `Irrelevant' and `non domain' unlike commonly used \textit{ordinal} grades. In the end-of-workshop report, Dzikovska et. al. discussed and compared submissions from 9 participating teams \cite{dzikovska2013semeval}.  The trend of feature design continued with most submissions \cite{Jimenez_softcardinality:hierarchical,celikouyekov13} employing various text similarity based features which were heavily tuned towards the dataset (with more emphasis on winning the task and less on generalizability). Multiple participants \cite{heilman2013ets,ott2013comet,zesch2013ukp} used some form of ``one-shot'' system combination approach, with several components feeding into a final decision made by a stacked classifier. One team later built on their submission and employed the idea of stacking on a reading comprehension dataset \cite{sakaguchi-heilman-madnani:2015:NAACL-HLT}. While our proposed ensemble-based technique also uses multiple classifiers, the gradual iterative transfer from the similarity based (second) classifier to the answer-text based (first) classifier makes it more robust as opposed to the existing one-shot stacking techniques (based on empirical evidence reported in Section~\ref{sec:results}). Kaggle, a platform for sharing data analytics problems and data, hosted a similar challenge to develop a scoring algorithm for short answers to 10 questions (reading comprehension and science) written by $10^{th}$ grade students.\footnote{\url{https://www.kaggle.com/c/asap-sas}} This dataset is one of the largest among public ASAG datasets in terms of number of students and also unique owing to the presence of well defined scoring schemes. Winning participants' reports, though not archived, again demonstrate prevalence of feature engineering with stacking of supervised methods.

We note that the features used for supervised ASAG techniques in different prior art are extensively tuned towards datasets used in respective papers. Rarely, a technique proposed in a paper is also tested on datasets referred to in prior papers. This lack of comparative analysis is also observed by both the recent survey papers \cite{burrows2015eras,roy2015perspective} who have independently emphasized the importance of sharing of data and ushered in the \textit{era of evolution} in ASAG. In this paper, we deliberately stayed away from dataset specific feature engineering for the second classifier (which depends on model answer) and rather used generic similarity measures at different types of textual representation (along the lines of unsupervised ASAG pioneered by \cite{mohler2009text}). Experimental results show that these measure work well across multiple datasets but we acknowledge that specific results may be improved by conducting focused dataset specific feature engineering (with explicit mentioning how it can be done in Section~\ref{sec:tech}).

\subsection{Transfer Learning}
Transfer learning \cite{tr_survey} in text analysis (a.k.a. domain adaptation\footnote{We use the terminologies \textit{transfer learning} and \textit{domain adaptation} interchangeably in this paper ignoring subtle, but irrelevant for our work, differences they bear.}) has shown promising results in recent years. \extra{Prior work on domain adaptation for text classification can be broadly classified into instance re-weighing and feature representation based adaptation approaches.}\fi A large body of domain adaptation literature are around  techniques which are based on learning common feature representation \cite{SCL1,daume2009frustratingly,\extra{jiang2007instance,}\fi SFA}. The intuitive idea behind most of these techniques is to learn a transformed feature space where if source and target domain instances are projected they follow a similar distribution. Consequently, a standard supervised learning algorithm can be trained on the former (projected source domain instances) to predict the latter (projected target domain instances). Structural Correspondence Learning (SCL) \cite{SCL1}, being one of the most widely used techniques, aims to learn the co-occurrence between
features expressing similar meaning in different
domains. In 2008, Pan et. al. \cite{pan2008transfer} proposed a dimensionality reduction method Maximum Mean Discrepancy Embedding to identify a shared latent space. \extra{Subsequently, they proposed to map domain specific words into unified clusters using
spectral clustering algorithm \cite{pan2008transfer}.}\fi \extra{In a follow up work, they demonstrated a novel feature
representation to perform domain adaptation
via Reproducing Kernel Hilbert Space using Maximum
Mean Discrepancy \cite{pan2011domain}.}\fi A similar approach,
based on co-clustering \cite{dhillon2003information} was
proposed by Dai et al. \cite{Tradaboost} to leverage common words as bridge between two domains. Daum\'{e} \cite{daume2009frustratingly} proposed a heuristic based non-linear mapping of source and target data to a high dimensional space. \extra{Bollegala
et al. \cite{Bollegala_kde} used sentiment sensitive thesaurus to
expand features for cross-domain sentiment classification.
In a comprehensive evaluation study, it
was observed that their approach tends to increase
the adaptation performance when multiple source
domains were used.\fi In this work, we used a classical feature mapping technique, canonical correlation analysis \cite{CCA2,CCA1}, towards learning a joint subspace where both the source and target domains features are mapped to have maximum correlation.

\extra{Instance re-weighing approaches address the
difference between the joint distributions of observed
instances and class labels in source domain
with that of target domain. Towards this direction, 
Liao et al. \cite{Liao} learned mismatch between
two domains and used active learning to
select instances from the source domain to enhance
adaptability of the classifier. Jiang and Zhai \cite{jiang_ser} proposed instance weighing scheme for domain
adaptation in NLP tasks which exploit independence
between feature mapping and instance
weighing approaches. Saha et al.\cite{Saha} leveraged
knowledge from source domain to actively
select the most informative samples from the target
domain. Xia et al. \cite{xia2013feature} proposed a hybrid
method for sentiment classification task that also
addresses the challenge of mutually opposite orientation
words.}\fi

\subsection{Transfer Learning and ASAG}
\label{sec:prior:tlasag}
Heilman and Madnani discussed about the use of domain adaptation for ASAG \cite{heilman2013ets} by applying the technique from \cite{daume2009frustratingly} to support generalization across questions and domains. For each feature, they maintained multiple copies with potentially different weights: a \textit{generic}
copy, a \textit{domain-specific} copy, and an \textit{item-specific} copy. For answers to a new question, only the
generic features get active but for answers
to questions in the training data, all copies
of the feature would be active and contribute to the
score. Apparently, for their submission in the SRA challenge, they used feature copying only for a subset of features. Phandi et. al. proposed a novel domain adaptation technique that uses Bayesian linear
ridge regression for a related task of automated essay scoring \cite{conf/emnlp/PhandiCN15}. Recently, Sultan et. al. \cite{SAG_NAACL} proposed a hierarchical Bayesian model for domain adaptation of short text where they mentioned possible application to short answer grading.

\section{Our Approach}
\label{sec:approach}
\subsection{Intuition}
In this section, we explain the proposed technique in an intuitive manner before describing the same formally in the next. Following transfer learning terminology, we refer to the questions for which graded answers are available as \textit{source questions} and questions for which no graded  answers are available as \textit{target questions}. Philosophy of our algorithm is gradual transfer of knowledge from a source to a target question while accounting for question specific variations. The technique has two salient features: an ensemble of two classifiers and an iterative transfer based on a common shared representation:

\textbf{Ensemble of classifiers}: We model ASAG as a supervised learning task where we employ an ensemble of two classifiers to predict student scores. 
In the ensemble, the first classifier is a text classifier trained on a bag of word (BoW) model of student answers. It is trained on the corpus of student answers only and does not require any model answer. The second classifier is based on real-valued features capturing similarity of student answers with respect to the model answer. While prior work have used many features towards capturing the same, we found often they were designed and tuned specifically for proprietary datasets. 
In our endeavor towards generalizability of the proposed technique, we employ five generic state of the art short-text similarity measures to compute similarity between the  model and student answers covering \textit{lexical}, \textit{semantic} and \textit{vector-space} measures. 
Additionally, the model of the first classifier is question specific (i.e. a word which is a good feature for a question is not necessarily a good feature for another question), whereas features for the second classifier are more question agnostic (i.e. high similarity with \textit{respective} model answer is indicative of high scores irrespective of question). The two classifiers thus capture complementary information useful for grading student answers. Finally, these two classifiers are combined in a weighted manner to form an ensemble which is used to predict the final score (label). 

\textbf{Transfer based on a common representation}: The ensemble of classifiers can be developed as described above for the source question based on instructor graded answers. The question is how do we do the same for target questions in absence of graded answers? It is done in two steps - (i) obtaining the second classifier through a feature based transfer of the model from the source to the target question, followed by (ii) iteratively building the first classifier and the ensemble using \textit{pseudo labeled data} from the target question. 

Learning a common representation for ASAG task is based on finding a common projection of the question agnostic features (used in the second classifier) for the  source and target questions. The common representation between the source and the target questions should be such that a model trained on this representation using graded student answers to the source question generalizes well for predicting the grades for the student answers to the target questions. For numeric features, we used the classical canonical correlation analysis (CCA) \cite{CCA2,CCA1} which aims to obtain a joint correlation subspace such that the projected features from the source and target domains are maximally correlated, as shown in Eq \ref{eq:cca}. Consider two random variables $X^s$ and $X^t$ such that $X^s=[x_1^s,....,x_n^s] \in R^{d_s\times n}$ and $X^t=[x_1^t,....,x_n^t] \in R^{d_t\times n}$. CCA is solved using generalized eigenvalue decomposition to obtain two projection vectors, 1) $p^s$ for the source and 2) $p^t$ for the target questions. 
\begin{equation}
\label{eq:cca}
\begin{split}
\max_{p^s,p^t} \rho =\frac{p^{s'} X^s X^{t'} p^t}{\sqrt[]{p^{s'} X^s X^{s'} p^s} \sqrt[]{p^{t'} X^t X^{t'} p^t}} \\ = \frac{p^{s'} \sum_{st} p^t}{\sqrt[]{p^{s'}\sum_{ss}p^s} \sqrt[]{{p^{t'}\sum_{tt}p^t}}}
\end{split}
\end{equation}
\noindent where $\sum_{tt}=X^tX^{t'}$, $\sum_{st}=X^sX^{t'}$, and $\sum_{sst}=X^sX^{s'}$ and $x'$ stands for transpose of $x$. Features extracted from the student answers to the source question are then projected onto this subspace (using $p^s$) to learn a model which is subsequently used to predict labels for the student answers to the target question projected onto the same subspace (using $p^t$).







The newly trained classifier on CCA-based transformed features is the second classifier of target question. It is applied to all student answers to the target question and \textit{confidently} predicted answers are chosen  as pseudo-labeled data to train the first classifier for the target question. We call this training data pool as pseudo as these are not labeled by the instructor rather based on (confident) predictions from the second classifier. The first classifier, trained on the text features using the pseudo labeled data, along with the transferred second classifier are combined as an ensemble (as described above) and applied on the remaining student answers to the target question (i.e. which were not in pseudo labeled training data). Confidently predicted instances from the ensemble are subsequently iteratively used to re-train the text classifier and boost up the overall prediction accuracy of the ensemble. The iteration continues till all the examples are correctly predicted or a specified number of iterations are performed.

\subsection{The Technique}
\label{sec:tech}
In this section, we describe the proposed technique considering two questions $q_s$ and $q_t$ as the \textit{source} and \textit{target} questions respectively. Notations used are shown in Table 2 and the block diagram depicting key steps is shown in Figure~\ref{fig:block}.

\begin{figure*}
\begin{center}
\fbox{\includegraphics[width=0.98\linewidth]{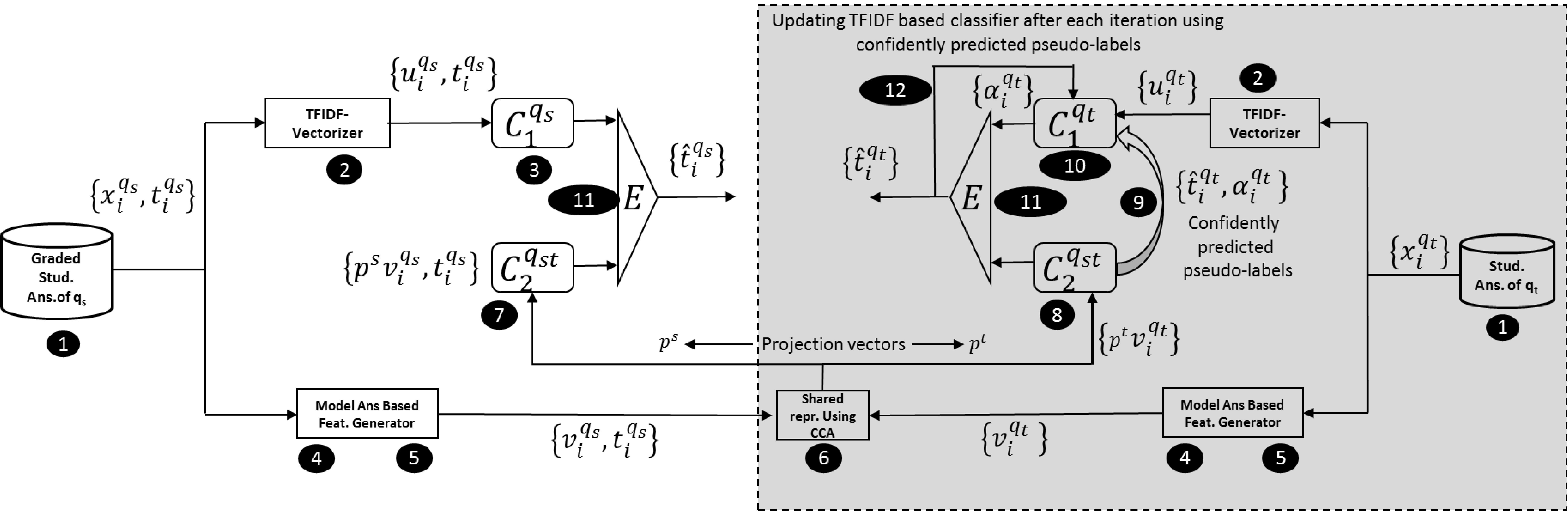}}
\caption{The block diagram of the proposed technique. The shaded part can be replicated for other target questions for which no labelled data is available. Numbers in black circle correspond to the step numbers in Section~\ref{sec:tech}.} 
\label{fig:block}
\end{center}
\end{figure*}

\begin{table}
\centering
\small
\label{tab:acc}
\begin{tabular}{|l|l|}
\hline
\multicolumn{1}{|c|}{\textbf{Symbol}} & \multicolumn{1}{|c|}{\textbf{Description}} \\ \hline
  $x_i^{q_j}$      &  $i^{th}$ student's answer to the $j^{th}$ question  \\ \hline
  $u_i^{q_j}$      &  TFDIF vector of bag of word representation of $x_i^{q_j}$  \\ \hline
  $v_i^{q_j}$      &    Vectors of features capturing similarity between \\
          &     student and model answers \\ \hline
  $t_i^{q_j}$      & Instructor given score of $x_i^{q_j}$, if available  \\ \hline
    $\hat{t}_i^{q_j}$      & Predicted score of $x_i^{q_j}$   \\ \hline
        $\alpha_i^{q_t}$     & Confidence of prediction for $i^{th}$ student answer to $q_t$   \\ \hline
  $p^s$, $p^t$ & CCA-based projection vectors for source \& target features \\\hline
  $C_1^{q_j}$      &  The first (text) classifier for ${q_j}$  \\ \hline
  $C_2^{q_{st}}$      &  The second classifier trained on CCA-based projections   \\ \hline
  $T$ & Pseudo-labeled training data in target \\\hline
  $E$ & Ensemble of $C_1^{q_j}$ and $C_2^{q_{st}}$  \\ \hline
  $\theta_1$, $\theta_2$  & Confidence threshold for $C_2^{q_{st}}$ and ensemble $E$ \\ \hline
 $w_1$, $w_2$  & Weights for ensemble $E$ \\ \hline
\end{tabular}
\label{tabnots}
\caption{List of notations used in this paper.}
\end{table}

\begin{enumerate}
\item Process graded student answers \{$x_i^{q_s}, t_i^{q_s}$\} of $q_s$ and ungraded answers \{$x_i^{q_t}$\} to create input vectors for two classifiers.
\item \textit{TFIDF-Vectorizer} for the graded answers of $q_s$ takes a bag-of-word (BoW) representations of student answers and converts to TFIDF vectors \{$u_i^{q_s}$\} with corresponding grades (labels), \{$t_i^{q_s}$\}. Prior to vectorization, we perform basic NLP pre-processing of stemming and stopword removal. We also perform question word demoting (i.e. considering words appearing in the question as stopwords while vectoring student answers) to avoid giving importance to parrot answering.
\item Train the first classifier $C_1^{q_s}$ on \{$u_i^{q_s}$, $t_i^{q_s}$\} using the graded answers of the source question ($q_s$).
\item Generate features for the second classifier using the following five similarity measures between student answers and model answer for a given question. All values are normalized between $0-1$ using min-max normalization leading to real valued vectors. Further, this classifier can be easily extended to include additional features that capture specific characteristics of the underlying dataset for enhanced performance. Many of such features are discussed in \cite{dzikovska2013semeval} (and the references there in); however, we restricted our proposed technique to general similarity based features rather than using features tailored for specific datasets.
\begin{itemize}
\item \textbf{LO}: First we consider lexical overlap (\textbf{LO}) between model and student answers. It is a simple baseline measure which looks for exact match for every content words (post pre-processing e.g. stopword removal and stemming) between student and model answers.
\item \textbf{JC and SP}: These are two semantic similarity measures based on Wordnet \cite{miller1995wordnet}. For each word in student answer, maximum word-to-word similarity scores are obtained with respect to words in model answers which are then summed up and normalized by the length of the two responses as described by Mohler and Mihalcea \cite{mohler2009text}. They compared eight options for computing word-to-word similarities; of which we select the two best performing ones viz. the measure proposed by Jiang and Conrath (\textbf{JC}) \cite{jiang1997semantic} and Shortest Path (\textbf{SP}).
\item \textbf{LSA and W2V}: These are the measures in vector space similarity category. In this category we first chose the most popular measure for measuring semantic similarity  viz. Latent Semantic Analysis (\textbf{LSA}) \cite{Landauer:1998} trained on a Wikipedia dump.  We also use the recently popular word2vec tool (\textbf{W2V}) \cite{mikolov2013efficient} to obtain vector representation of words which  are trained on 100 billion words of Google news dataset and are of length 300. Word-to-word similarity measures obtained using euclidean distance between word vectors are summed up and normalized in a manner similar to \textit{JC} and \textit{SP}. 
\end{itemize}

\item  Compute \{$v_i^{q_s}, t_i^{q_s}$\} for the $i^{th}$ student answer to source question $q_s$, and  \{$v_i^{q_t}$\} for the $i^{th}$ student answer to the target question $q_t$.
\item Learn CCA-based projection vectors $p^s$ and $p^t$ to transform the real valued features from the source and target questions respectively to have maximum correlation, as shown in Eq \ref{eq:cca}. 

\item Train $C_2^{q_{st}}$ using CCA transformed features on the graded answers to the source question, \{$p^sv_i^{q_s}, t_i^{q_s}$\}.

\item Use $C_2^{q_{st}}$ to predict labels, $\hat{t}_i^{q_t}$, of student answers to $q_t$ on the CCA-based representation $p^tv_i^{q_t}$.

\item Move instances which are predicted with confidence greater than a pre-defined threshold, $\theta_1$,  to a pseudo training data-pool $T$. 
\item TFIDF vectorized representation of instances in $T$ are selected to train the first classifier (text based) $C_1^{q_t}$ for the target question (Same as Steps 2 \& 3 for $q_s$). 
\item The two classifiers $C_1^{q_t}$ \& $C_2^{q_{st}}$ are combined to form an ensemble $E(\cdot) \to w_1 C_1^{q_t}(\cdot) + w_2 C_2^{q_{st}}(\cdot)$. The ensemble $E$ is used to predict the remaining instances and the instances now predicted with a confidence greater than another predefined threshold $\theta_2$ are again added to $T$.
\item Update/re-train the first classifier $C_1^{q_t}$ using additional pseudo-labeled instances (added in the previous step).
\item \textbf{Step-13:} Update ensemble weights based on the error of individual classifiers such that the better classifier gets more weight mass.
\vspace{-6pt}
\begin{equation}
\label{eq:weights}
w_1^{l+1}=1-\frac{w_1^{l}*I(C_1^{q_t})}{w_1^{l}*I(C_1^{q_t})+w_2^{l}*I(C_2^{q_t})};w_2^{l+1}=1-w_1^{l+1}
\end{equation}

\vspace{-6pt}

\noindent $I(\cdot)$ is absolute error of individual classifiers w.r.t. to the pseudo labels obtained by the ensemble over all confidently predicted instances in $l^{th}$ iteration.
\item Repeat steps $10$ to $12$  till all instances are confidently predicted or a specified number of iterations are performed.
\end{enumerate}

Algorithm \ref{alg:online} summarizes the proposed algorithm for automatic short answer grading. The iterative algorithm converges when no more student answers to the target question can be confidently predicted or maximum number of iterations are performed ($iterMAX=10$ in our experiments). The transfer of knowledge occurs within the ensemble where the first classifier trained on the CCA-based shared representation provides pseudo-labeled training data to initialize the first classifier on the TFIDF representations of student answers. These two complementary classifiers are further combined in an ensemble where the CCA-based classifier helps in better learning the first classifier. Finally, the ensemble is used for predicting the labels for the ungraded student answers for enhanced ASAG performance. The underlying classifiers used in the proposed algorithm in the logistic regression (LR) classifier and the weights for the individual classifiers in the ensemble are initialized to $0.5$ and are further updated as shown in Eq. \ref{eq:weights}. The thresholds $\theta_1$ \& $\theta_2$ are empirically set to $2$ / (\textit{number~of~class-labels}) where \textit{number~of~class-labels} $>2$ for all questions. 

An intrigued reader may find the proposed iterative approach similar to the classical co-training algorithm proposed by Blum and Mitchell \cite{Blum:1998:CLU:279943.279962}. 
However, there are significant differences between the proposed iterative technique and co-training. Firstly, in co-training the assumption is that one has labeled data, albeit in small number, according to both views of the data. In the proposed technique we do not require \textit{any} labeled data for the target question. In fact, the proposed technique can be deployed for a large number of target questions requiring graded answers from only a few source questions. Secondly, in co-training both classifiers get updated in an iterative manner whereas in the proposed technique only the text classifier gets updated.


\begin{algorithm}[t]
   \caption{\textbf{The Proposed Algorithm for ASAG}}
   \label{alg:online}
\begin{algorithmic}
   \STATE {\bfseries INPUT:} $C_2^{q_{st}}$ trained using $\{p^s\textbf{v}_i^{q_s},t_i^{q_s}\}$ from $q_s$, thresholds $\theta_1$ \& $\theta_2$, $T=\emptyset$, $n$: count of student answers to $q_t$.
\vspace{4pt}

\STATE{\bfseries 1: INITIALIZE TFIDF-based CLASSIFIER in TARGET:\\}  

  \FOR{$i=1$ {\bfseries to} $n$}
  \bindent
   \STATE $ C_2^{q_{st}}\left(p^t v_i^{q_t}\right) \to \hat{t}_i^{q_t}$ \& calculate confidence of prediction ($\alpha_i$).
  
  \IF{$\alpha_i > \theta_1$}
   \STATE Move $x_i^{q_t}$ to $T$ with pseudo label as $\hat{t}_i^{q_t}$.
   \ENDIF{.}
      \eindent
   \ENDFOR

\STATE Initialize $C_1^{q_t}$ using pseudo-labeled instances in $T$

\vspace{4pt}

\STATE{\bfseries 2: ITERATIVE LEARNING:\\}  

  \STATE{\bfseries Iterate:} $l = 0$ : till $l\leq$ $iterMax$  
   \STATE {\bfseries Process:} Construct $E (\cdot) \to w_1 C_1^{q_t}(\cdot) + w_2 C_2^{q_{st}}(\cdot)$ 
   \FOR{$i=1$ {\bfseries to} $n-||T||$}
   \STATE Predict labels: $E\left(p^t v_i^{q_t}\right) \to \hat{t_i}^{q_t}$; calculate $\alpha_i$: confidence of prediction.
   
  \IF{$\alpha_i > \theta_2$}
   \STATE Add $x_i^{q_t}$ to $T$ with pseudo label $\hat{t}_i^{q_t}$.
   \ENDIF{.}
   \ENDFOR{\bfseries.} \\   
   Retrain $C_1^{q_t}$ on $T$ \& update ensemble weights.
   \STATE{\bfseries end iterate.}
   \STATE {\bfseries OUTPUT:} Updated TFIDF-based classifier $C_1^{q_t}$.
  \end{algorithmic}
\end{algorithm}

\section{Experimental Evaluation}
\label{sec:eval}
\subsection{Datasets}
Prior work in (supervised) ASAG has not presented evaluation results on multiple datasets. In fact, the recent survey papers referred to in Section~\ref{sec:prior} (\cite{burrows2015eras,roy2015perspective}) have emphasized the  need for sharing of datasets and structured evaluations of techniques across multiple datasets. Towards that, we have evaluated the proposed technique on four datasets covering varying subject matter (sciences and literature) as well as standards (high school and college). 

{\bf SE2013:}\footnote{\url{https://www.cs.york.ac.uk/semeval-2013/task7/index.php\%3Fid=data.html}} This dataset is a part of the ``Student Response Analysis'' (SRA) in the Semantic Evaluation (SemEval) workshop in 2013~\cite{dzikovska2013semeval}. The task released two datasets: BEETLE data, based on transcripts of students interacting with BEETLE II tutorial dialogue system~\cite{conf/acl/DzikovskaMSCFC10}, and S\textsc{ci}E\textsc{nts}B\textsc{ank} data based on the corpus of student answers to assessment questions collected by~\cite{nielsen2008annotating}. For this work, we only consider S\textsc{ci}E\textsc{nts}B\textsc{ank} dataset as it had exactly one model answer for every question. Our technique could be extended in future for BEETLE
dataset as well where questions have varying number
of model answers.  The S\textsc{ci}E\textsc{nts}B\textsc{ank}  training corpus contains approximately 10,000 answers to 197 assessment questions from 15 different science domains. The answers were graded by multiple annotators on a nominal scale viz. `Correct', `Partially correct incomplete',`Contradictory' (student answer contradicts the reference answer), `Irrelevant' and `non domain'. Test subsets are of three types:
\begin{itemize}
\item \textbf{Unseen answers (UA)}: A held-out set of student answers to the questions contained in the training set.
\item \textbf{Unseen questions (UQ)}: A test set created by holding back all student answers to a subset of randomly selected questions. These questions were not present in the training set but they are from the same domain.
\item \textbf{Unseen domains (UD)}: Same as UQ but test set questions are from different domains than training set. 
\end{itemize}

{\bf CSD:}\footnote{\url{http://web.eecs.umich.edu/~mihalcea/downloads/ShortAnswerGrading_v1.0.tar.gz}} This is one of the earliest ASAG datasets comprising of a set of questions, model answers and student answers taken from an undergraduate computer science course \cite{mohler2009text}. The data set consists of 21 questions (7 questions from 3 assignments each) from introductory assignments in the course with answers provided by a class of abut 30 undergraduate students. Student answers were independently evaluated by two annotators on a scale of 0-5 and automatic techniques are measured against their average. All our detail analysis are reported based on this dataset. 

{\bf X-CSD:}\footnote{\url{http://web.eecs.umich.edu/~mihalcea/downloads/ShortAnswerGrading_v2.0.zip}} This is an extended version of CSD with 87 questions from the same course \cite{mohler2011learning}.

{\bf RCD:} We created a new dataset on  a reading comprehension assignment for Standard-12 students in a Central Board of Secondary Education (CBSE) school in India. The dataset contains 14 questions based on a passage which were answered by 58 students. The answers were graded by two human raters based on model answers and an optional scoring scheme.

\subsection{Performance Metrics}
Depending on the nature of labels (grades) i.e. ordinal or nominal, we used two different performance metrics. Most ASAG datasets (in our case CSD, X-CSD, and RCD) have ordinal class labels; hence we used \textit{mean absolute error} (MAE) as the metric for quantitative evaluation. MAE for a question is the absolute difference between the groundtruth and predicted scores averaged over all students and  is given by $\frac{1}{n}\sum_{i=1}^n|t_i-y_i|$,
\extra{
\begin{equation}
MAE = \frac{1}{n}\sum_{i=1}^n|t_i-y_i|
\end{equation}
}\fi
where $t_i$ and $y_i$ are respectively the groundtruth and predicted scores of the $i^{th}$ student's answer. For reporting aggregate performances over a  dataset, question wise MAE values are averaged for all questions.

The SE2013 dataset has nominal class labels. Following the evaluation metrics used in the SRA task, we report two confusion matrix based evaluation metrics viz. the \textit{macro-average} $F_1$ (= $1/N_c \sum_c F_1(c)$) and \textit{weighted average} $F_1$ (= $1/N \sum_c\  |c| \times  F_1(c)$) as described in the end-of-workshop report \cite{dzikovska2013semeval}. Here $N_c$ is the number of
classes (e.g. `correct', `contradictory' etc.), $N$ is the
total number of test items, $|c|$ is the number of items labeled as $c$ in gold-standard data and $F_1(c)$ is class specific $F_1$ score for class $c$.
As described in \cite{dzikovska2013semeval}, we ignore the `nondomain' class as it is severely underrepresented and report macro-averaged $F_1$ over 4 classes for consistent comparison.

\extra{ and $F_1$ is defined as below:
\begin{equation}
F_1 = \frac{1}{|C|}\sum_{c=1}^{|C|}\frac{2P_cR_c}{P_c+R_c}
\end{equation}
where $P_c$ and $R_c$ are precision and recall for class $c$. As described in the original paper\cite{dzikovska2013semeval}}\fi

\subsection{Quantitative Results}
\label{sec:results}
In this section, we first present aggregate results for all the datasets followed by fine grained  results and insights on the CSD dataset.
\subsubsection{Aggregate Results}
\label{sec:agg_results}
Table~\ref{tab:semeval} shows performance of the proposed technique on SE2013 dataset against the entry ``ETS''   \cite{heilman2013ets} (the only ASAG technique based on transfer learning as reviewed in Section~\ref{sec:prior:tlasag}) as well as the best performances obtained for the SRA task in SemEval workshop reported in \cite{dzikovska2013semeval}. We report two runs of ``ETS'' as their performances varied significantly between them. The proposed technique performs better than ``ETS'' as well as the best performing entries across \textbf{all} test sets in terms of both the metrics. It is important to note that all techniques use labeled data in supervised learning mode whereas the proposed technique requires labeled data only from the source question. This is a significant feature of the proposed technique, demonstrating that using labeled data from the source question along with generic similarity measures between the student and model answers can result in efficient ASAG in the target question without any labeled data.\footnote{While we focused on the finer 5-way categorization, we observed similar results for 2-way and 3-way tasks in SRA (created by combining `Partially correct incomplete', `Irrelevant', and `non domain' (and `Contradictory' for 2-way)).}

\begin{table}[]
\centering
\begin{tabular}{|l|l|l|l||l|l|l|}
\hline
                                                                               & \multicolumn{3}{c||}{\textbf{Weighted average $F_1$}}                                                         & \multicolumn{3}{c|}{\textbf{Macro-average $F_1$}}                                                    \\ \hline
                                                                               & \multicolumn{1}{c|}{\textbf{UA}} & \multicolumn{1}{c|}{\textbf{UQ}} & \multicolumn{1}{c||}{\textbf{UD}} & \multicolumn{1}{c|}{\textbf{UA}} & \multicolumn{1}{c|}{\textbf{UQ}} & \multicolumn{1}{c|}{\textbf{UD}} \\ \hline
\textbf{ETS$_{\text{1}}$}                                                                  & 0.535                            & 0.487                            & 0.447                            & 0.467                            & 0.372                            & 0.334                            \\ \hline
\textbf{ETS$_{\text{2}}$}                                                                  & 0.625                            & 0.356                            & 0.434                            & 0.581                            & 0.274                            & 0.339                            \\ \hline
\textbf{\begin{tabular}[c]{@{}l@{}}Best \\ Performance \\ in SRA\end{tabular}} & 0.625                            & 0.492                            & 0.471                            & 0.581                            & 0.384                            & 0.375                            \\ \hline
\textbf{Proposed}                                                              & \textbf{0.672}                   & \textbf{0.518}                   & \textbf{0.507}                   & \textbf{0.612}                                & \textbf{0.415}                                & \textbf{0.402}                                \\ \hline
\end{tabular}
\caption{Comparison of $F_1$ scores (higher the better) of the proposed technique against the transfer learning based ASAG technique (ETS$_{\text{1}}$ and ETS$_{\text{2}}$) \cite{heilman2013ets} and the best performance obtained in the SRA task. Existing results are from \cite{dzikovska2013semeval}.}
\label{tab:semeval}
\end{table}
For the other three datasets with ordinal labels, there was no prior work based on transfer learning approach to ASAG. We followed the convention in transfer learning literature of comparing against a skyline and a baseline:
\begin{itemize}
\item {\bf Baseline (Sup-BL)}: Supervised models are built using labeled data from a source question and applied \textit{as-it-is} to a target question.
\item {\bf Skyline (Sup-SL)}: Supervised models are built assuming labeled data is available for all questions (including target). Performance is measured by training a model on every question and applied on the same.
\end{itemize}
Performance of transfer learning techniques should be in between the baseline and skyline - closer to the skyline, better it is. As shown in Table~\ref{tab:summary}, the proposed method beats the baseline for all datasets handsomely (differences being $0.35$, $0.03$ and $0.86$ for CSD, X-CSD, and RCD respectively) whereas coming much closer to the skyline (differences being $0.01$, $0.28$ and $0.05$ for CSD, X-CSD, and RCD respectively).\footnote{We exclude questions which do not have short answers viz. questions marked with \# sign by authors in X-CSD and questions $6$ and $11$ in RCD.}

\begin{table}
\centering
\begin{tabular}{|l|l|l|l|}
\hline
                  & \textbf{CSD} & \textbf{X-CSD} & \textbf{RCD} \\ \hline
\textbf{SUP-BL}   & 0.95         & 0.85           & 1.74     \\ \hline
\textbf{SUP-SL}   & 0.66         & 0.54           & 0.83             \\ \hline
\textbf{Proposed} & 0.67         & 0.82           & 0.88            \\ \hline 
\end{tabular}
\caption{Overall performance (MAE; lower the better) of the proposed technique along with the baseline and the skyline performances on the three data sets. Sup-SL requires labelled data for all questions unlike Sup-BL and the proposed technique.}
\label{tab:summary}
\end{table}

\subsubsection{Detailed Results}
\label{sec:detail_results}



\begin{table}
\centering
\small
\label{tab:CSD_acc}
\begin{tabular}{|c|c|c|c|c|}
\hline
\multirow{2}{*}{\textbf{Question}} & \multirow{2}{*}{\textbf{Sup-BL}} & \textbf{CCA-based } & \multirow{2}{*}{\textbf{Ensemble}} & \multirow{2}{*}{\textbf{Sup-SL}} \\
&&\textbf{classifier ($C_2$)}&&\\\hline
1  & 1.57  &1.10  & 0.52  & 1.55    \\ \hline
2  & 1.59   &0.48  & 0.48  & 1.27    \\ \hline
3  & 1.18   &0.81  & 0.81  & 0.66    \\ \hline
4  & 0.59   &0.16  & 0.16  & 0.50    \\ \hline
5  & 0.72   &0.19  & 0.19  & 0.59    \\ \hline
6  & 1.31   &0.55  & 0.55  & 1.17    \\ \hline
7  & 0.71   &2.48  & 2.23  & 0.58    \\ \hline
8  & 0.74   &0.90  & 0.84  & 0.63    \\ \hline
9  & 0.90   &0.35  & 0.35  & 0.53    \\ \hline
10 & 1.09   &1.19  & 1.23  & 0.47    \\ \hline
11 & 0.70   &0.74  & 0.74  & 0.07    \\ \hline
12 & 1.70   &1.52  & 1.52  & 1.53    \\ \hline
13 & 0.78   &0.26  & 0.26  & 0.52    \\ \hline
14 & 1.29   &0.65  & 0.65  & 0.90    \\ \hline
15 & 0.61   &0.23  &0.23   & 0.33    \\ \hline
16 & 0.78   &0.35   & 0.35  & 0.27    \\ \hline
17 & 0.76   &1.32   & 1.00   & 0.70    \\ \hline
18 & 0.91   &0.48  & 0.48  & 1.16    \\ \hline
19 & 0.87   &1.03  & 1.03  & 0.47    \\ \hline
20 & 0.59   &0.39   & 0.39  & 0.26    \\ \hline
21 & 0.62   &0.23   & 0.23  & 0.57    \\ \hline
                                  
\end{tabular}
\label{tab:qwise_result}
\caption{Question wise performance comparison of the MAE (lower the better) achieved by different algorithms on source question selected using the minimum proxy $\mathcal{A}$-distance on the CSD dataset.}
\end{table}

Most prior work in supervised ASAG has reported only aggregated performances over all questions. However, we note that performance of ASAG techniques varies significantly across questions as well as depends on other factors such as the choice of classifier or source question. In this section, we present detailed results with explanation and insights. For lack of space, we present the detailed results only for the CSD dataset.

{\bf Question-wise performance:}  Table V \cb{hardcoded table number}\fi compares the question-wise MAE of the proposed algorithm against the skyline and baseline on all the $21$  questions in the CSD dataset. For Sup-BL and the proposed technique, for each question we consider all remaining 20 questions as {\it source} one at a time and report the best MAE obtained. Firstly, we note that all methods exhibit performance variations across questions. Variance of SUP-BL, CCA classifier, SUP-SL and the proposed technique are $0.12$, $4.46$, $0.16$ and $0.24$ respectively. Secondly, for all questions the proposed technique gives MAE between those of Sup-BL and Sup-SL while being closer to the Sup-SL as observed in aggregate result (Table~\ref{tab:summary}).
Thirdly, the proposed algorithm which combines the text and numeric classifier in a weighted ensemble yields significantly lower error rates than the constituent classifiers. This demonstrates the fact that the ensemble exploits their complimentary nature effectively towards improving the overall performance.

\extra{ {\bf Variation with source question:}  Figure \ref{fig:res_csd} shows variation in performance of the proposed algorithm on all the $21$ questions in the CSD dataset. The box plot shows variation in performance on each question  when the transfer is performed using the remaining questions ~(one-at-a-time). The box plot for questions $3$, $4$, and $12$ is comparatively shorter than the remaining ones which suggest that the overall transfer performance from all the remaining questions has a high level of agreement with each other. On the other hand, the box plots for questions $1$, $2$, $16$, $20$, and $21$ are comparatively tall which suggest that the transfer performances from different questions are quite varied. 
Questions $2$ and $6$ have almost the same median MAE; however, the box plots for these questions show very different distributions of the transfer error rates. 
}\fi

\extra{
\begin{figure}[t]
\begin{center}
\fbox{\includegraphics[width=0.98\linewidth]{boxplot.jpg}}
\caption{Question wise variation in performance of the proposed algorithm on the CSD dataset. The bar ranges from minimum MAE to the maximum MAE on a target question with the median MAE highlighted with a horizontal black line.} \label{fig:res_csd}
\end{center}
\end{figure}
}\fi

                                  

\begin{figure}[t]
\begin{center}
\fbox{\includegraphics[width=0.98\linewidth]{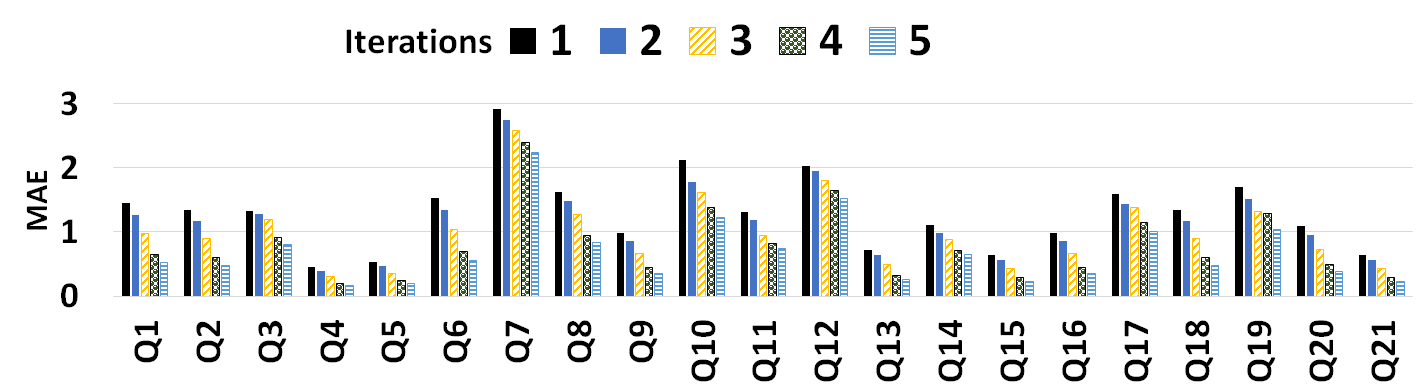}}
\caption{Effect on the average performance of the ensemble when the TFIDF-based classifier is progressively updated with pseudo-labeled instances obtained at each iteration. The graph shows the drop in MAE during the first $5$ iterations of the proposed algorithm on the CSD dataset.} \label{fig:iterations}
\end{center}
\end{figure}

\textbf{Effect of Iterative Learning:}  Figure \ref{fig:iterations} shows the effect of iteratively building the first (text) classifier for the target question based on the pseudo-labeled training instances provided by the second (numeric) classifier. Results suggest that the iterative learning monotonically reduces MAE on all the questions. This further validates our assertion that exploiting textual features along with features derived from  model answers in an iterative manner is an important aspect in ASAG. We observed that for most questions the iterative algorithm converged in $5-6$ iterations.

\begin{table}
\centering
\small
\label{tab:CSD_classifiers}
\begin{tabular}{|c|c|c|c|c|c|}
\hline
\textbf{Question} & \textbf{LDA} & \textbf{Adaboost}  & \textbf{SVM} & \textbf{LR} & \textbf{Q-wise}\\
 &  &   &  &  & \textbf{Variance}\\\hline
  Q1 &1.38&1.26& 0.93&0.81 & 0.12\\\hline
 Q2 &1.42&1.31& 0.86&0.73&0.16\\\hline 
 Q3 &2.03&1.94& 1.62&1.24&0.21\\\hline
 Q4 &1.20&1.06& 0.78&0.57&0.13\\\hline
 Q5 &1.14&0.98& 0.70&0.48&0.24\\\hline
 Q6 &1.28&1.10& 0.82&0.94&0.21\\\hline
 Q7 &3.06&3.10& 2.94&2.69&0.26\\\hline
 Q8 &1.72&1.68& 1.49&1.37&0.17\\\hline
 Q9 &1.40&1.44& 1.12&0.96&0.20\\\hline
 Q10 &2.25&2.32& 2.11&1.72&0.23\\\hline  
 Q11 &1.70&1.75& 1.32&1.10&0.09\\\hline 
 Q12 &2.26&2.34& 2.04&1.84&0.26\\\hline 
 Q13 &1.30&1.27& 0.85&0.51&0.19\\\hline
 Q14 &1.19&1.16& 1.02&0.87&0.11\\\hline
 Q15 &1.58&1.63& 0.88&0.58&0.27\\\hline
 Q16 &1.52&1.42& 0.94&0.62&0.17\\\hline
 Q17 &1.80&1.71& 1.96&1.78&0.19\\\hline
 Q18 &1.56&1.53& 1.07&0.81&0.22\\\hline
 Q19 &1.91&1.83& 1.84&1.68&0.14\\\hline
 Q20 &1.22&1.15& 0.98&0.73&0.17\\\hline
 Q21 &1.44&1.38& 0.85&0.64&0.23\\\hline
 \textbf{Mean MAE}&\textbf{1.63}&\textbf{1.58}& \textbf{1.29}&\textbf{1.07}&\\\hline
                                  
\end{tabular}
\label{tab:diff_classifiers}
\caption{Question wise MAE of the proposed technique with different classifiers on CSD. Numbers reported are the average MAEs based on transfer from all the remaining $20$ questions.}
\end{table}

\textbf{Effect of Different Classifiers:} To explore if the proposed technique generalizes across different classifiers, we experimented with multiple classifiers to build the ensemble viz. logistic regression~(LR), support vector machines ~(SVM), AdaBoost and linear discriminant analysis ~(LDA). Table VI \cb{hardcoded table number}\fi shows that the performance of the proposed algorithm with respect to underlying classifiers. 
It is observed that performance is slightly better with LR  (Mean MAE=1.07) and SVM (1.29) as compared to LDA (1.63) and Adaboost (1.58). Another implication of this result is that one can tune the second classifier with more number of dataset specific features (as reported in Section~\ref{sec:prior}) and still be able to use the proposed technique as a framework. While we deliberately avoided such feature engineering in this paper, it would be an interesting study as a future work.

\textbf{Effect of Different Grading Schemes:} Each question in the CSD dataset are graded in the range $0-5$ leading to $11$ possible scores ($0,0.5, \ldots, 4.5, 5$). Hence this is a $11$-class classification problem with only $30$ student answers as dataset. Even with a leave-one-out experimental protocol this is an extremely sparse training dataset with most classes having no training examples. We analyze the effect of different grading schemes on the performance of the proposed algorithm under three different granularity of grading schemes: 1) $11$-class grading scheme ranging from $0$-$5$ with a step size of $0.5$, 2) $6$-class grading scheme ranging from $0$-$5$ with a step size of $1$ and 3) $2$-class grading scheme with score $>3$ as correct and $\leq3$ as incorrect. Results in Figure \ref{fig:grading} reports the performance of the proposed algorithm with the three grading schemes and it suggests that the MAE reduce get better with coarser scoring schemes.  \extra{For higher number of classes, i.e. 11-class grading scheme, the number of training instances per class are sparse. This sparsity reduces as the classes are merged from 11-class to 6-class to 2-class grading scheme and hence leads to enhances ASAG performance.}\fi  While such reduction in MAE with coarser class labels is expected, we believe that with larger amount of student data we will be able to further reduce MAE even at finer class label structure.

\begin{figure}
\begin{center}
\fbox{\includegraphics[width=0.9\linewidth]{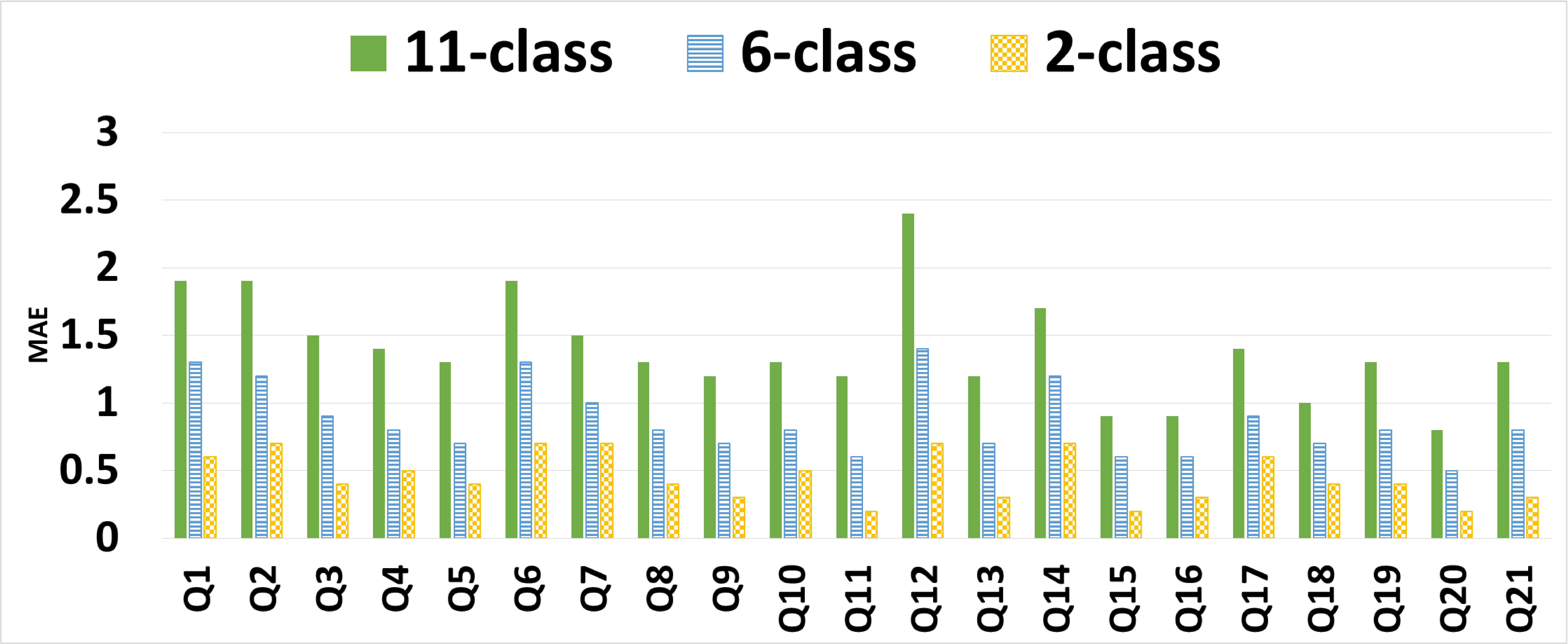}}
\caption{Performance of the proposed algorithm with different grading schemes (ranging from $11$-to-$6$-to-$2$ class problem) on the CSD dataset.} \label{fig:grading}
\end{center}
\end{figure}

                                  




\section{Conclusion}
\label{sec:conc}
In this paper, we presented a novel ASAG technique based on an ensemble of a text and a numeric classifier of complementary nature. The technique used a canonical correlation analysis based transfer learning to bootstrap an iterative algorithm to obtain the ensemble for target questions without requiring \textit{any} labeled data. We demonstrated efficacy of the proposed technique by empirical evaluation on multiple datasets from different subject matters and standards. In future, we intend to conduct studies towards comparing various feature representation techniques along with prior art in supervised ASAG. Additionally, it will be interesting to compare deep learning techniques against heavy feature engineering based approaches prevalent in supervised ASAG prior art. Another interesting question that came up during the course of this work, is that certain types of questions are perhaps more amenable to be recipient of transfer than others. If yes, then how do we characterize those based on questions and model answers? Finally, through this work, we introduced the potential of application of transfer learning in supervised ASAG towards making it practical which hopefully would bring in more novel work in this direction.
\extra{
\section{Acknowledgment}
We acknowledge our \textit{kind colleague-1} for experimental help towards feature generation based on similarity measures.
}\fi





%

\bibliographystyle{IEEEtran}
\bibliography{edu}
\end{document}